%% file: main.tex
\documentclass[letterpaper, 10 pt, conference]{ieeeconf}  

\IEEEoverridecommandlockouts                              

\usepackage{multicol}
\usepackage{graphicx}
\usepackage{amsmath}
\usepackage{bbm}
\usepackage{dsfont}
\usepackage{dblfloatfix}
\usepackage[ruled]{algorithm2e}

\SetCommentSty{mycommfont}
\SetFuncSty{myprocname}

\newcommand\blfootnote[1]{%
  \begingroup
  \renewcommand\thefootnote{}\footnote{#1}%
  \addtocounter{footnote}{-1}%
  \endgroup
}

\usepackage{mathrsfs}
\usepackage{bm}
\usepackage[makeroom]{cancel}
\usepackage{amssymb}
\usepackage{upgreek}
\usepackage{multirow}
\usepackage{amsfonts}
\usepackage{here}
\usepackage[normalem]{ulem}
\usepackage{mdframed}

\usepackage{booktabs}
\usepackage{color}
\usepackage{subcaption}
\usepackage[font=small, skip=5pt]{caption}
\usepackage{listings}
\usepackage{hyperref}
\hypersetup{
    colorlinks,
    linkcolor={black},
    citecolor={blue!80!black},
    urlcolor={blue!80!black}
}
\usepackage{wrapfig}
\usepackage{pdflscape}

\usepackage{cleveref}
\usepackage{tikz}

\definecolor{mycolor1}{HTML}{000080}
\definecolor{mycolor2}{HTML}{AB1117}
\definecolor{mycolor3}{HTML}{006413}
\definecolor{mycolor4}{HTML}{EC6505}
\definecolor{mycolor5}{HTML}{24AFF0}
\definecolor{mycolor6}{HTML}{FF7034}
\definecolor{mycolor7}{HTML}{00CED1}
\definecolor{mycolor8}{HTML}{FACE30}
\definecolor{mycolor9}{HTML}{BCBCBC}

\newcommand{\Sspace}{\mathcal{S}}

\newcommand{\Ospace}{\Omega}
\newcommand{\E}{\mathbb{E}}

\newcommand{\abs}[1]{\lvert#1\rvert}

\newcommand{\sprl}{\mathcal{R}}

\makeatletter
\newcommand*\bigcdot{\mathpalette\bigcdot@{.5}}
\newcommand*\bigcdot@[2]{\mathbin{\vcenter{\hbox{\scalebox{#2}{$\m@th#1\bullet$}}}}}
\makeatother

\makeatletter
\let\NAT@parse\undefined
\makeatother
\usepackage[numbers]{natbib}

\title{\LARGE \bf
  Spatial Language Understanding for Object Search in Partially Observed City-scale Environments
}

\author{Kaiyu Zheng, Deniz Bayazit, Rebecca Mathew, Ellie Pavlick, Stefanie Tellex\\
  Department of Computer Science, Brown University\\
  \small{\texttt{\{kzheng10, dbayazit, stefie10\}@cs.brown.edu, \{rebecca\_mathew, ellie\_pavlick\}@brown.edu}}}

\begin{document}

\maketitle
\thispagestyle{empty}
\pagestyle{empty}

\begin{abstract}
  Humans use spatial language to naturally describe object locations and their relations. Interpreting spatial language not only adds a perceptual modality for robots, but also reduces the barrier of interfacing with humans. 
  Previous work primarily considers spatial language as goal specification for instruction following tasks in fully observable domains, often paired with reference paths for reward-based learning. However, spatial language is inherently subjective and potentially ambiguous or misleading. Hence, in this paper, we consider spatial language as a form of stochastic observation. We propose SLOOP (Spatial Language Object-Oriented POMDP), a new framework for partially observable decision making with a probabilistic observation model for spatial language. We apply SLOOP to object search in city-scale environments. To interpret ambiguous, context-dependent prepositions (e.g.~\emph{front}), we design a simple convolutional neural network that predicts the language provider's latent frame of reference (FoR) given the environment context. Search strategies are computed via an online POMDP planner based on Monte Carlo Tree Search. Evaluation based on crowdsourced language data, collected over areas of five cities in OpenStreetMap, shows that our approach achieves faster search and higher success rate compared to baselines, with a wider margin as the spatial language becomes more complex. Finally, we demonstrate the proposed method in AirSim, a realistic simulator where a drone is tasked to find cars in a neighborhood environment.
\end{abstract}

\section{Introduction}
\label{sec:intro}
\blfootnote{Project website: \url{https://h2r.github.io/sloop/}}
Consider the scenario in which a tourist is looking for an ice cream truck in an amusement park. She asks a passer-by and gets the reply \emph{the ice cream truck is behind the ticket booth}. The tourist looks at the amusement park map and locates the ticket booth. Then, she is able to infer a region corresponding to that statement and find the ice cream truck, even though the spatial preposition \emph{behind} is inherently ambiguous and subjective to the passer-by. Robots capable of understanding spatial language can leverage prior knowledge possessed by humans to search for objects more efficiently, and interface with humans more naturally. Such capabilities can be useful for applications such as autonomous delivery and search-and-rescue, where the customer or people at the scene communicate with the robot via natural language.

This problem is challenging because humans produce diverse spatial language phrases based on their observation of the environment and knowledge of target locations, yet none of these factors are available to the robot. In addition, the robot may operate in a different area than where it was trained. The robot must generalize its ability to understand spatial language across environments.

Prior works on spatial language understanding assume referenced objects already exist in the robot's world model~\cite{tellex2011forklift, Fasola2013UsingSS,janner2018representation} or within the robot's field of view~\cite{Blukis:18drone}. Works that consider partial observability do not handle ambiguous spatial prepositions \cite{hemachandra2015learning,wandzel2019oopomdp} or assume a robot-centric frame of reference \cite{bisk2018learning,patki2020language}, limiting the ability to understand diverse spatial relations that provide critical disambiguating information, such as \emph{behind the ticket booth}.
For downstream tasks, existing works primarily consider spatial language as goal or trajectory specification \cite{vogel-jurafsky-2010-learning,Kollar2010TowardUNInstructionFollowing}. They require large datasets of spatial language paired with reference paths to learn a policy by end-to-end reward-based learning \cite{sotp2019acl,wang2019reinforced} or imitation learning \cite{blukis2018mapping}, which is expensive to acquire for realistic environments (such as urban areas), and generalization to such environments is an ongoing challenge \cite{Blukis:18drone,bisk2016natural,blukis2019learning}.

\begin{figure}
    \centering
    \includegraphics[width=\linewidth]{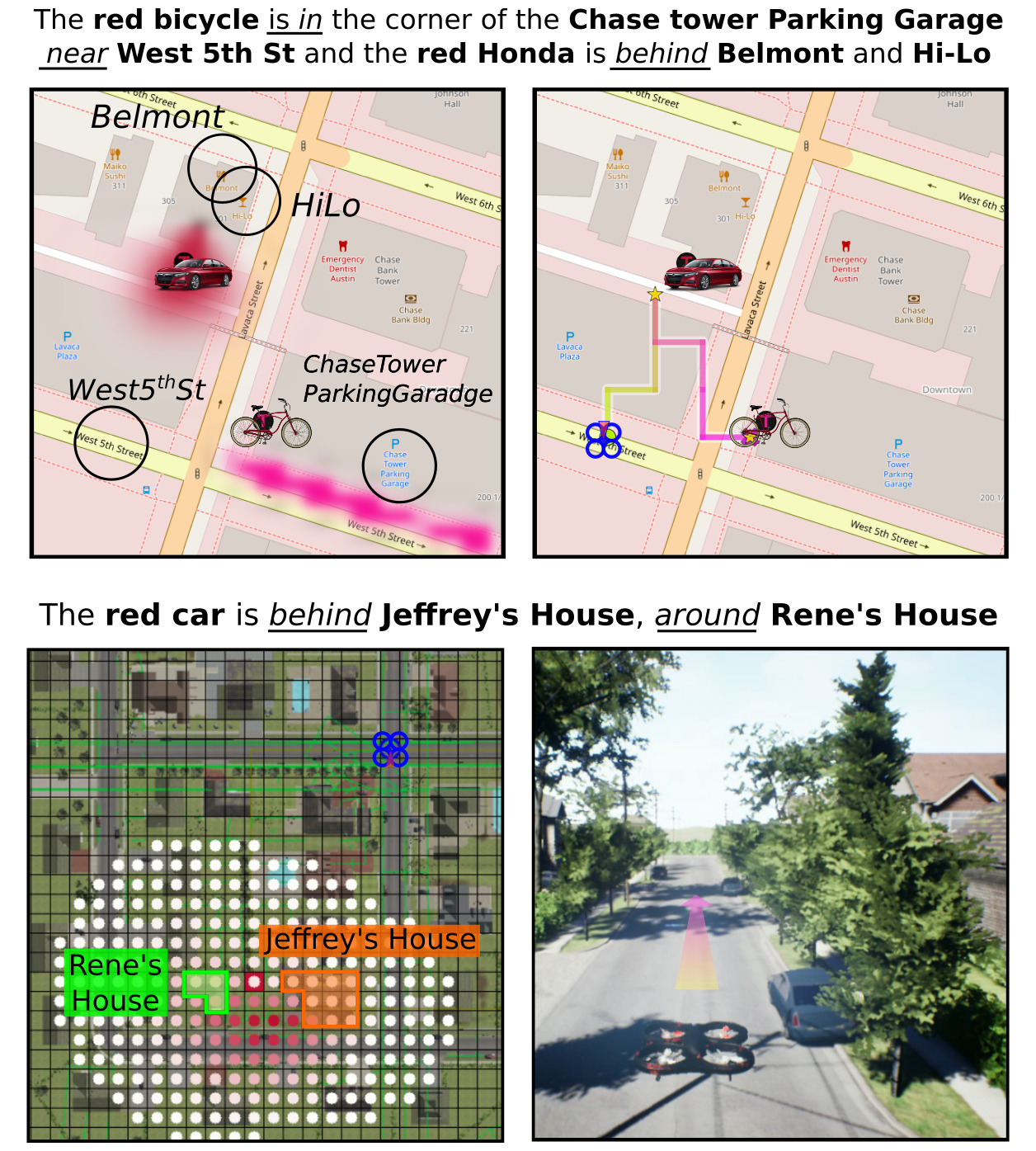}
    \caption{
      Given a spatial language description, a drone with limited field of view must find target objects in a city-scale environment. Top row: example trial from OpenStreetMap \cite{OpenStreetMap}. Bottom row: example trial from AirSim \cite{airsim2017fsr}. Left side: belief over target location after incorporating spatial language using the proposed approach. Right side: Screenshot of simulation with search path.
    }
    \vspace{-0.5cm}
    \label{fig:firstfig}
  \end{figure}

  Partially Observable Markov Decision Process (POMDP) \cite{KAELBLING199899} is a principled decision making framework widely used in the object search literature \cite{aydemir2013active,xiao2019online,zheng2020multi}, due to its ability to capture uncertainty in object locations and the robot's perception. \citet{wandzel2019oopomdp} proposed Object-Oriented POMDP (OO-POMDP), which factors the state and observation spaces by objects and is designed to model tasks in human environments. In this paper, we introduce SLOOP (Spatial Language Object-Oriented POMDP), which extends OO-POMDP by considering spatial language as an additional perceptual modality. We derive a probabilistic model to capture the uncertainty of the language through referenced objects and landmarks. This enables the robot to incorporate into its belief state spatial information about the referenced object via belief update.

  We apply SLOOP to object search in city-scale environments given a spatial language description of target locations. We collected a dataset of five city maps from OpenStreetMap \cite{OpenStreetMap} as well as spatial language descriptions through Amazon Mechanical Turk (AMT). To understand ambiguous, context-dependent prepositions (e.g.~\emph{behind}), we develop a simple convolutional neural network that infers the latent frame of reference (FoR) given an egocentric synthetic image of the referenced landmark and surrounding context. This FoR prediction model is integrated into the spatial language observation model in SLOOP.  We apply POMCP \cite{silver2010monte}, an online planning algorithm based on Monte-Carlo Tree Search (MCTS) to plan search behavior, after the initial belief update upon receiving the spatial language. Note that in general, because SLOOP regards spatial language as an observation, the language can be received during task execution.

We evaluate both the FoR prediction model and the object search performance under SLOOP using the collected dataset.
Results show that our method leads to search strategies that find objects faster with higher success rate by exploiting spatial information from language compared to a keyword-based baseline used in prior work \cite{wandzel2019oopomdp}. We also report results for varying language complexity and spatial prepositions to discuss advantages and limitations of our approach.
Finally, we demonstrate SLOOP for object search in AirSim \cite{airsim2017fsr}, a realistic drone simulator
shown in Fig.~\ref{fig:firstfig}, where the drone is tasked to find cars in a neighborhood environment. 

\begin{figure}[t]
    \centering
    \includegraphics[width=\linewidth]{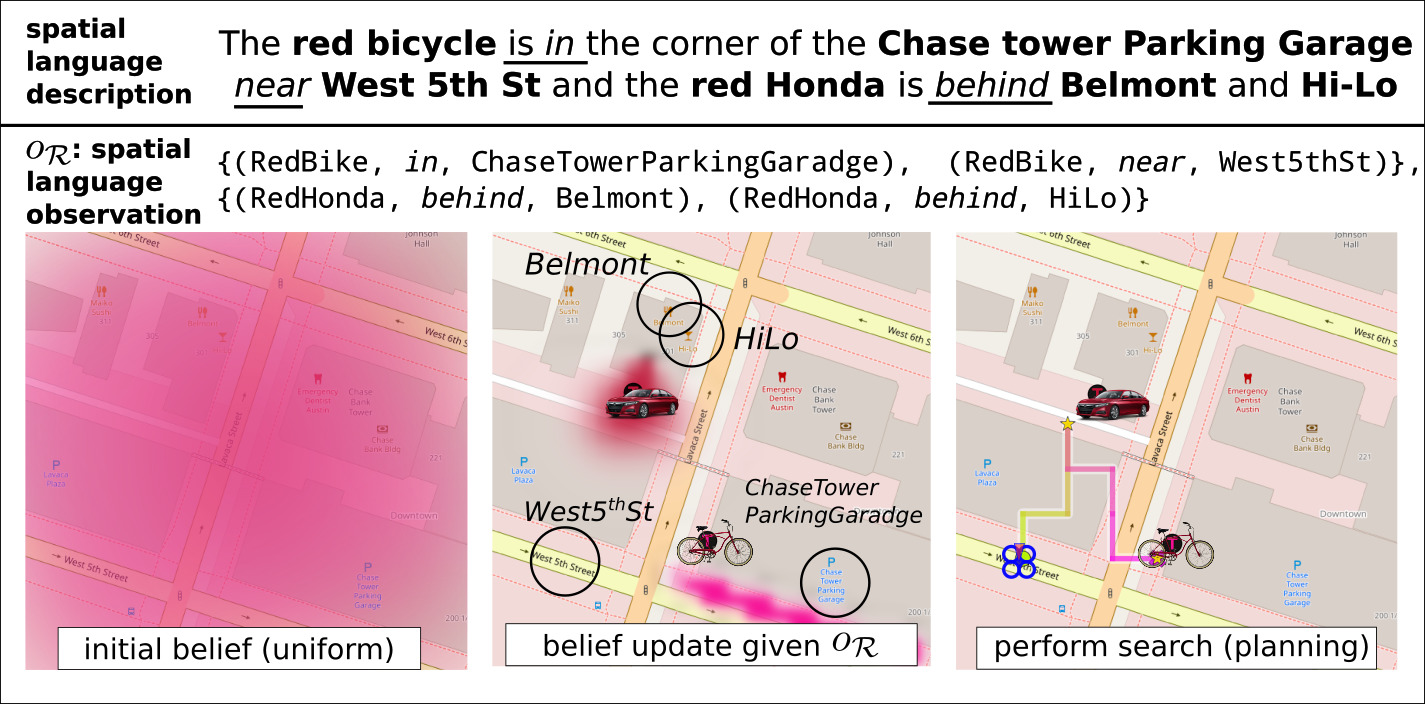}
    \caption{
      We consider a spatial language description as an observation $o_{\sprl}$, which is a set of $(f, r, \gamma)$ tuples, obtained through parsing the input language. We propose an observation model that incorporates the spatial information in $o_{\sprl}$ into the robot's belief about target locations, which benefits subsequent object search performance.
    }
    \vspace{-0.5cm}
    \label{fig:method}
  \end{figure}

\section{Related Work}
\label{sec:related_work}
Spatial language is a way of communicating spatial information of objects and their relations using spatial prepositions (e.g. \emph{on}, \emph{between}, \emph{front}) \cite{hayward1995spatial}. Understanding spatial language for decision making requires the robot to map symbols of the given language description to concepts and structures in the world, a problem referred to as language grounding. Most prior works on spatial language grounding assume a fully observable domain \cite{tellex2011forklift, Fasola2013UsingSS, Blukis:18drone, vogel-jurafsky-2010-learning, Kollar2010TowardUNInstructionFollowing}, where the referenced objects have known locations or are within the field of view, and the concern is synthesizing navigation behavior faithful to a given instruction (e.g. \emph{Go to the right  side of the rock} \cite{Blukis:18drone}). Recent works aim to map such instructions directly to low-level controls or navigation trajectories leveraging deep reinforcement learning \cite{vogel-jurafsky-2010-learning,sotp2019acl,Blukis2020:fewshot-drone} or imitation learning \cite{Blukis:18drone, wang2019reinforced,blukis2018mapping}, requiring large datasets of instructions paired with demonstrations. In this work, the referenced target objects have unknown locations, and the robot has a limited field of view. We regard the spatial language description as observation and obtain policy through online planning.

Spatial language understanding in partially observable environments is an emerging area of study \cite{hemachandra2015learning, wandzel2019oopomdp, patki2020language, thomason2019visiondialogue}. \citet{thomason2019visiondialogue} propose a domain where the robot, tasked to reach a goal room, has access to a dialogue with an oracle discussing the location of the goal during execution. \citet{hemachandra2015learning} and \citet{patki2020language} infer a distribution over semantic maps for instruction following then plan actions through behavior inference. These instructions are typically FoR-independent or involve only the robot's own FoR. In contrast, we consider language descriptions with FoRs relative to referenced landmarks. \citet{wandzel2019oopomdp} propose the Object-Oriented POMDP framework for object search and a proof-of-concept keyword-based model for language understanding in indoor space. Our work handles diverse spatial language using a novel spatial language observation model and focuses on search in cityscale environments. We evaluate our system against a keyword-based baseline similar to the one in \cite{wandzel2019oopomdp}.

Cognitive scientists have grouped FoRs into three categories: absolute, intrinsic, and relative \cite{majid2004can,shusterman2016frames}. Absolute FoRs (e.g. for \emph{north}) are fixed and depend on the agreement between speakers of the same language. Intrinsic FoRs (e.g. for \emph{at the door of the house}) depend only on properties of the referenced object.  Relative FoRs (e.g. for \emph{behind the ticket booth}) depend on both properties of the referenced object and the perspective of the observer. In this paper, spatial descriptions are provided by human observers who may impose relative FoRs or absolute FoRs.

\section{Preliminaries: POMDP and OO-POMDP}
\label{sec:prelim}
POMDP \cite{KAELBLING199899} is a framework to describe sequential-decision making problems where the agent does not fully observe the environment state. Formally, a POMDP is defined as a tuple $\langle S, A, \Ospace, T, O, R, \delta\rangle$.
Given an action $a\in A$, the environment state transitions from $s\in\Sspace$ to $s'\in S$ following the transition function $T(s,a,s')=\Pr(s'|s,a)$. The agent receives an observation  $o\in\Ospace$ according to the observation model $O(s',a,o)=\Pr(o|s',a)$, and an expected reward $r=R(s,a,s')$. The agent maintains a \emph{belief state} $b_t(s) = \Pr(s|h_t)$ which is a sufficient statistic for the history $h_t=(ao)_{1:t-1}$. The agent updates its belief given the action and observation by $b_{t+1}(s')=\eta O(s',a,o)\sum_s T(s,a,s')b_t(s)$ where $\eta={\sum_s\sum_{s'}\Pr(o|s',a)\Pr(s'|s,a)b_t(s)}$ is the normalizing constant. The task of the agent is to find a policy $\pi(b_t)\in A$ which maximizes the
expectation of future discounted rewards ${V^{\pi}}(b_t)=\E\left[\sum_{k=0}^{\infty}\delta^{k}R(s_{t+k},\pi(b_{t+k}), s_{t+k+1})\ |\ b_t\right]$ 
with a discount factor
$\delta$.

An Object-Oriented POMDP (OO-POMDP) \cite{wandzel2019oopomdp} (generalization of OO-MDP
  \cite{diuk2008object}) is a POMDP that considers the state
  and observation spaces to be factored by a set of $n$ objects, $\Sspace = \Sspace_1\times \cdots \times \Sspace_n$, $\Ospace = \Ospace_1\times\cdots\times \Ospace_n$, where each object belongs to a class with a set of attributes. A simplifying assumption is made for the 2D MOS domain \cite{wandzel2019oopomdp} that objects are independent so that the belief space scales linearly rather than exponentially in the number of objects: $b_t(s)=\prod_ib_t^i(s_i)$.

\section{Problem Formulation}
\label{sec:setting}
We are interested in the problem setting similar to the opening scenario in the \nameref{sec:intro}. A robot is tasked to search for $N$ targets in an urban or suburban area represented as a discrete 2D map of landmarks $\mathcal{M}$. The robot is equipped with the map and can detect the targets within the field of view.  However, the robot has no knowledge of object locations a priori, and the map size is substantially larger than the sensor's field of view, making brute-force search infeasible. A human with access to the same map and prior knowledge of object locations provides the robot with a natural language description. For example, given the map in Fig.~\ref{fig:firstfig}, one could say \emph{the red car is behind Jeffrey's House, around Rene's House.} The language is assumed to mention some of the target objects and their spatial relationships to some of the known landmarks, yet there is no assumption about how such information is described.
To be successful, the robot must incorporate information from spatial language to efficiently search for the target object.

This problem can be formulated as the multi-object search (MOS) task \cite{wandzel2019oopomdp}, modeled as an OO-POMDP. The state $s_i=(x_i,y_i)$ is the location for target $i$, $1\leq i\leq N$. The robot state $s_r=(x,y,\theta,\mathcal{F})$ consists of the robot's pose $(x,y,\theta)$ and a set of found targets $\mathcal{F}\subseteq\{1,\cdots,N\}$. There are three types of actions: \textsc{Move} changes the robot pose (possibly stochastically); \textsc{Look} processes sensory information within the current field-of-view; \textsc{Find}($i$) marks object $i$ as found. In our implementation of MOS, a \textsc{Look} action is automatically taken following every \textsc{Move}. Upon taking \textsc{Find}, the robot receives reward $R_{\text{max}}\gg 0$ if an unfound object is within the field of view, and $R_{\text{min}}\ll 0$ otherwise. Other actions receive $R_{\text{step}}<0$. The desired policy accounts for the belief over target locations while efficiently exploring the map.

Note that in our evaluation, we use a synthetic detector that returns observations conditioned on the ground truth object locations for belief update during execution. Our POMDP-based framework can easily make use of a realistic observation model instead, for example, based on processing visual data \cite{xiao2019online, monso2012pomdp}. Training vision-based object detectors is outside the scope of this paper. Our focus is on spatial language understanding for planning object search strategies.

\section{Technical Approach}


In this section, we introduce SLOOP, with a particular focus on the observation space and observation model. Then, to apply SLOOP to our problem setting, we describe our implementation of the spatial language observation model on 2D city maps, which includes a convolutional network model for FoR prediction.


\subsection{Spatial Language Object-Oriented POMDP (SLOOP)}

SLOOP augments an OO-POMDP defined over a given map $\mathcal{M}$ with a spatial language observation space and a spatial language observation model. The map $\mathcal{M}$ consists of a discrete set of locations and contains landmark information (e.g. landmark's name, location and geometry), such that $N$ objects exist on the map at possibly unknown locations. Thus, the state space can be factored into a set of $N$ objects plus the given map $\mathcal{M}$ and robot state $s=(s_1,\cdots,s_N,s_r,\mathcal{M})$. SLOOP does not augment the action space, thus the action space of the underlying OO-POMDP is left unchanged. Because the transition and reward functions are, by definition, independent of observations, they are also kept unchanged in SLOOP. Next, we introduce spatial language observations.

\label{sec:integrate_pomdp}

According to \citet{landau1993whatwhere}, the standard linguistic representation of an object's place requires three elements: the object to be located (\emph{figure}), the reference object (\emph{ground}), and their relationship (\emph{spatial relation}). We follow this convention and represent spatial information from a given natural language description in terms of atomic propositions, each represented as a tuple of the form $(f,r,\gamma)$, where $f$ is the \emph{figure}, $r$ is the \emph{spatial relation} and $\gamma$ is the \emph{ground}. 
In our case, $f$ refers to a target object, $\gamma$ refers to a landmark on the map, and $r$ is a predicate that is true if the locations of $f$ and $\gamma$ satisfy the semantics of the spatial relation.
As an example, given spatial language \emph{the red Honda is behind Belmont, near Hi-Lo}, two tuples of this form can be extracted (parentheses indicate role): \{(\texttt{RedCar}($f$), \texttt{behind}($r$), \texttt{Belmont}($\gamma$)), (\texttt{RedCar}($f$), \texttt{near}($r$), \texttt{HiLo}($\gamma$))\}.

We define a \emph{spatial language observation} as a set of $(f,r,\gamma)$ tuples extracted from a given spatial language. We define the spatial language observation space to be the space of all possible tuples of such form for a given map, objects, and a set of spatial relations.

We denote a spatial language observation as $o_{\sprl}$. Our goal now is to derive $\Pr(o_{\sprl}|s',a)$, the observation model for spatial language. We can split $o_{\sprl}$ into subsets, $o_{\sprl}=\cup_{i=1}^No_{\sprl_i}$, where each $o_{\sprl_i}=\cup_{k=1}^{L}(f_i,r_k,\gamma_k)$, $L=|o_{\sprl_i}|$ is the set of tuples where the figure is target object $i$. Since the human describes the target objects with respect to landmarks on the map, the spatial language is conditionally independent from the robot state and action given map $\mathcal{M}$ and the target locations $s_1,\cdots,s_N$. Therefore,
\begin{align}
    \Pr(o_{\sprl}|s',a)&=\Pr(\cup_{i=1}^N o_{\sprl_i}|s_1',\cdots,s_N',\mathcal{M})
\end{align}
We make a simplifying assumption that $o_{\sprl_i}$ is conditionally independent of all other $o_{\sprl_j}$ and $s_{j}'$ ($j\neq i$) given  $s_i'$ and map $\mathcal{M}$. We make this assumption because the human observer is capable of producing a language $o_{\sprl_i}$ to describe target $i$ given just the target location $s_i$ and the map $\mathcal{M}$. Thus,
\begin{align}
  \Pr(\cup_{i=1}^N o_{\sprl_i}|s_1',\cdots,s_N',\mathcal{M})=\prod_{i=1}^N\Pr(o_{\sprl_i} | s_i', \mathcal{M})
\end{align}
where $\Pr(o_{\sprl_i} | s_i', \mathcal{M})$ models the spatial information in the language description for object $i$. For each spatial relation $r_j$ in $o_{\sprl_i}$ whose interpretation depends on the FoR imposed by the human observer (e.g. \emph{behind}), we introduce a corresponding random variable $\Psi_j$ denoting the FoR vector that distributes according to the indicator function $\Pr(\Psi_j=\psi_j) = \mathds{1}(\psi_{j}=\psi_{j}^*)$, where $\psi_j^*$ is the one imposed by the human, unknown to the robot. Then, our model for $\Pr(o_{\sprl_i} | s_i', \mathcal{M})$ becomes:
\begin{align}
  \Pr(o_{\sprl_i} | s_i', \mathcal{M}) =\prod_{j=1}^L \Pr(r_j | \gamma_j, \psi_j^*, f_i, s_i', \mathcal{M}) \label{eq:sprls}
\end{align}
The step-by-step derivation can be found in the supplementary material. It is straightforward to extend this model as a mixture model $\Pr(o_{\sprl_i} | s_i', \mathcal{M})=\sum_{k=1}^mw_k\Pr_k(o_{\sprl_i} | s_i', \mathcal{M})$, $\sum_{k=1}^mw_k=1$, where multiple interpretations of the spatial language are used to form separate distributions then combined into a weighted-sum. This effectively smooths the distribution under individual interpretations, which improves object search performance in our evaluation (Figure~\ref{fig:detections}),

However, to proceed modeling $\Pr(r_j | \gamma_j, \psi_j^*, f_i, s_i', \mathcal{M})$ in Eq.~(\ref{eq:sprls}), we notice that it depends on the unknown FoR $\psi_j^*$. Therefore, we consider two subproblems instead: The approximation of $\psi_j^*$ by a predicted value $\hat{\psi}$ and the modeling of $\Pr(r_j | \gamma_j, \hat{\psi_j}, f_i, s_i', \mathcal{M})$. Next, we describe our approach to these two subproblems for object search in city-scale environments.

\subsection{Learning to Predict Latent Frame of Reference}
\label{sec:for_prediction}

\begin{figure*}[t]
  \centering
  \includegraphics[width=0.75\linewidth]{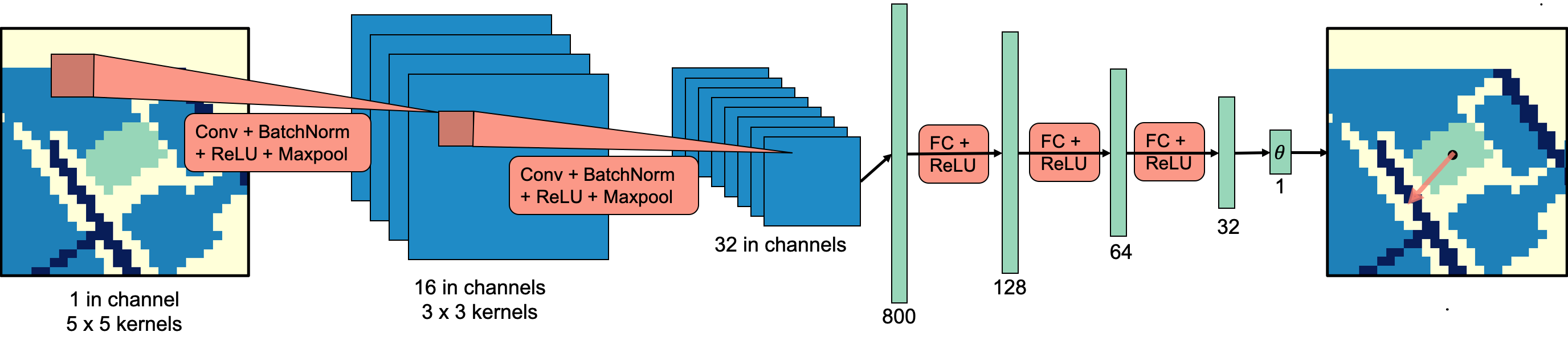}
  \caption{\textbf{Frame of Reference Prediction Model Design.} In this example taken from our dataset, the model is predicting the frame of reference for the preposition \emph{front}. The grayscale image is rendered with color ranging from blue (black), green (gray) to yellow (white). Green highlights the referenced landmark, dark blue the streets, blue the surrounding buildings, and yellow the background.}
  \label{fig:model_design}
\end{figure*}
Here we describe our approach to predict $\psi_j^*$ corresponding to a given $(f_i,r_j,\gamma_j)$ tuple, which is critical for correct resolution of spatial relations. Taking inspiration from the ice cream truck example where the tourist can infer a potential FoR by looking at the 2D map of the park, we train a model that predicts the human observer's imposed FoR based on the environment context embedded in the map.

We define an FoR in a 2D map as a single vector $\psi_{j}=(x,y,\theta)$ located at $(x,y)$ at an angle $\theta$ with respect to the $+x$ axis of the map.
We use the center-of-mass of the ground as the origin $(x,y)$. We make this approximation since our data collection shows that when looking at a 2D map, human observers tend to consider the landmark as a whole without decomposing it into parts. Therefore, the FoR prediction problem becomes a regression problem of predicting the angle $\theta$ given a representation of the environment context.


We design a convolutional neural network, which
takes as input a grayscale image representation of the environment context where the ground in the spatial relation is highlighted. Surrounding landmarks are also highlighted with different brightness for streets and buildings (Figure~\ref{fig:model_design}). The image is shifted to be egocentric with respect to the referenced landmark, and cropped into a 28$\times$28 pixel image. The intuition is to have the model focus on immediate surroundings, as landmarks that are far away tend not to contribute to inferring the referenced landmark's properties. The model consists of two standard convolution modules followed by three fully connected layers.
These convolution modules extract an 800-dimension feature vector, feeding into the fully connected layers, which eventually output a single value for the FoR angle. We name this model \textbf{EGO-CTX} for egocentric shift of the contextual representation.

Regarding the loss function,
a direct comparison with the labeled FoR angle is not desirable. For example, suppose the labeled angle is $0$ (in radians). Then, a predicted angle of $0.5$ is qualitatively the same as another predicted angle of $-0.5$ or $2\pi - 0.5$.
For this reason, we apply the following treatment to the difference between predicted angle $\theta$ and the labeled angle $\theta^*$. Here, both $\theta$ and $\theta^*$ have been reduced to be between $0$ to $2\pi$:
\begin{align}
\label{eq:lossf}
\ell(\theta,\theta^*)=
    \begin{cases}
      2\pi - \abs{\theta-\theta^*}, & \text{if}\  \abs{\theta-\theta^*} > \pi,\\
      \abs{\theta-\theta^*}, & \text{otherwise}\\
    \end{cases}
\end{align}
This ensures that the angular deviation used to compute the loss ranges between $0$ to $\pi$.
The learning objective is to reduce such deviation to zero. To this end, we minimize the mean-squared error loss $L(\bm{\theta},\bm{\theta}^*) = \frac{1}{N}\sum_{i=1}^N\left( \ell (\theta_i,\theta^*_i)\right)^2$,
where $\bm{\theta},\bm{\theta}^*$ are predicted and annotated angles in the training set of size $N$. This objective gives greater penalty to angular deviations farther away from zero.

In our experiments, we combine the data by antonyms and train two models for each baseline: a \textbf{front} model used to predict FoRs for \emph{front} and \emph{behind}, and a \textbf{left} model used for \emph{left} and \emph{right}\footnote{We do not train a single model for all four prepositions since \emph{left} and \emph{right} often also suggest an \emph{absolute} FoR used by the language provider when looking at 2D maps, while \emph{front} and \emph{behind} typically suggest a \emph{relative} FoR.}.
We augment the training data by random rotations for \textbf{front} but not  for \textbf{left}.\footnote{Again, because \emph{left} and \emph{right} may imply either absolute or relative FoR.} We use the Adam optimizer~\cite{kingma2014adam} to update the network weights with a fixed learning rate of $1\times 10^{-5}$. Early stopping based on validation set loss is used with a patience of 20 epochs~\cite{prechelt1998early}.


\subsection{Modeling Spatial Relations}


We model $\Pr(r_j | \gamma_j, \hat{\psi_j}, f_i, s_i', \mathcal{M})$ as a Gaussian following prior work and evidence from animal behavior \cite{fasola2013iros, o1996geometric}:
\begin{align}
\begin{split}
&\Pr(r_j | \gamma_j,\hat{\psi}_j,f_i, s_i', \mathcal{M})\\
&\qquad=\abs{u(s_i',\gamma_j,\mathcal{M})\bigcdot v(f_i,r_j,\gamma_{j},\hat{\psi}_{j})}\\
&\qquad\qquad\times\exp\left(-dist(s_i',\gamma_j,\mathcal{M})^2/2\sigma^2\right)
\end{split}
\end{align}
where $\sigma$ controls the steepness of the distribution based on the spatial relation's semantics and landmark size, and $dist(s_i',\gamma_j,\mathcal{M})$ is the distance between $s_i'$ to the closest position within the ground $\gamma_j$ in map $\mathcal{M}$, and $u(s_i',\gamma_j,\mathcal{M})\bigcdot v(f_i,r_j,\gamma_{j},\hat{\psi}_{j})$ is the dot product between $u(s_i',\gamma_j,\mathcal{M})$, the unit vector from $s_i'$ to the closest position within the ground $\gamma_j$ in map $\mathcal{M}$, and $v(f_i,r_j,\gamma_{j},\hat{\psi}_{j})$, a unit vector in in the direction that satisfies the semantics of the proposition $(f_i,r_j,\gamma_j)$ by rotating $\hat{\psi}_{j}$.
The dot product is skipped for prepositions that do not require FoRs (e.g. \emph{near}). We refer to \citet{landau1993whatwhere} for a list of prepositions meaningful in 2D that require FoRs: \emph{above, below, down, top, under, north, east, south, west, northwest, northeast, southwest, southeast, front, behind, left, right}.

\section{Data Collection}
\label{sec:data_collect}
In this section, we describe our data collection process as well as a pipeline for spatial information extraction from natural language.
We use maps from OpenStreetMap (OSM), a free and open-source database of the world map with voluntary landmark contributions \cite{OpenStreetMap}.
We scrape landmarks in 40,000m$^2$ grid-regions with a resolution of 5m by 5m grid cells in five different cities leading to a dimension of 41$\times$41 per grid map\footnote{Because of the curvature of the earth, the grid cells and overall region is not perfectly square, which is why the grid is not perfectly 40x40}: Austin, TX; Cleveland, OH; Denver, CO; Honolulu, HI, and Washington, DC. Geographic coordinates of OSM landmarks are translated into grid map coordinates.


To collect a variety of spatial language descriptions from each city, we randomly generate 30 environment configurations for each city, each with two target objects.
We prompt Amazon Mechanical Turk (AMT) workers to describe the location of the target objects and specify that the robot knows the map but does not know target locations.
Each configuration is used to obtain language descriptions from up to eleven different workers.
The descriptions are parsed using our pipeline described next in Sec.~\ref{sec:spatial_info_extraction}. Examples are shown in Fig.~\ref{fig:data_info}. Screenshots of the survey and statistics of the dataset are provided in the supplementary material.

The authors annotated FoRs for \emph{front, behind, left} and \emph{right} through a custom annotation tool which displays the AMT worker's language alongside the map without targets.
We manually infer the FoR used by the AMT worker, similar to what the robot is tasked to do.
This set of annotations are used as data to train and evaluate our FoR prediction model. Prepositions such as \emph{north, northeast} have absolute FoRs with known direction. Others are either difficult to annotate (e.g. \emph{across}) or have too little samples (e.g. \emph{above}, \emph{below}).

\subsection{Spatial Information Extraction from Natural Language}
\label{sec:spatial_info_extraction}


We designed a pipeline to extract spatial relation triplets from the natural language using the spaCy library \cite{spaCy2} for noun phrase (NP) identification and dependency parsing, as it achieves good performance on these tasks. Extracted NPs are matched against synonyms of target and landmark symbols using cosine similarity. All paths from targets to landmarks in the dependency parse tree are extracted to form the $(f,r,\gamma)$ tuples used as spatial language observations (Sec.~\ref{sec:integrate_pomdp}).

Our spatial language understanding models assume as input language that has been
parsed into $(f,r,\gamma)$ tuples, but is not dependent on this exact pipeline for
doing so. Future work could explore alternative methods for parsing and entity
linking, including approaches optimized for the task of spatial language
resolution. In our end-to-end experiments, we report the task performance both when using this parsing pipeline and when using manually annotated $(f,r,\gamma)$ tuples to indicate the influence of parsing on search performance.


\begin{figure}[t]
\centering
\includegraphics[width=\linewidth]{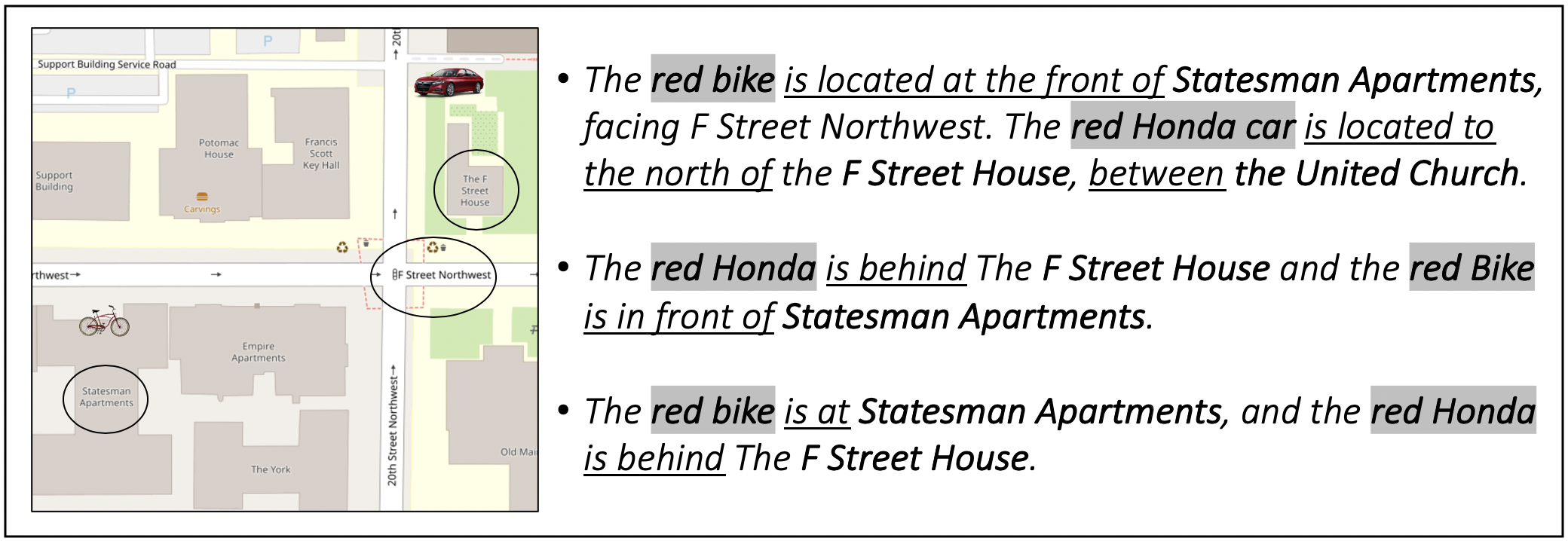}
\caption{Map screenshot shown to AMT workers paired with collected spatial language descriptions.}
\label{fig:data_info}
\end{figure}


\section{Experiments}
\label{sec:evaluation}


\subsection{Frame of Reference Prediction}
\label{sec:foref_eval}
We test the generalizability of our FoR prediction model (\textbf{EGO-CTX}) by cross-validation. The model is trained on maps from four cities and tested on the remaining held-out city for all possible splits. We evaluate the model by the angular deviation between predicted and annotated FoR angles, in comparison with three baselines and human performance:
The first is a variation (\textbf{CTX}) that uses a synthetic image with the same kind of contextual representation
yet without egocentric shift. The second is another variation (\textbf{EGO}) that performs egocentric shift and also crops a $28\times 28$ window, but only highlights the referenced landmark at the center without contextual information. The random baseline (\textbf{Random}) predicts the angle at random uniformly between $[0,2\pi]$. The \textbf{Human} performance is obtained by first computing the differences between pairs of annotated FoR angles for the same landmarks (Eq.~\ref{eq:lossf}), then averaging over all such differences for landmarks per city.
Each pair of FoRs may be annotated by the same or different annotators.
Taking the average gives a sense of the disagreement among the annotators' interpretation of spatial relations.

\begin{figure}[t]
\centering
\vspace{-0.3cm}
  \includegraphics[width=\linewidth]{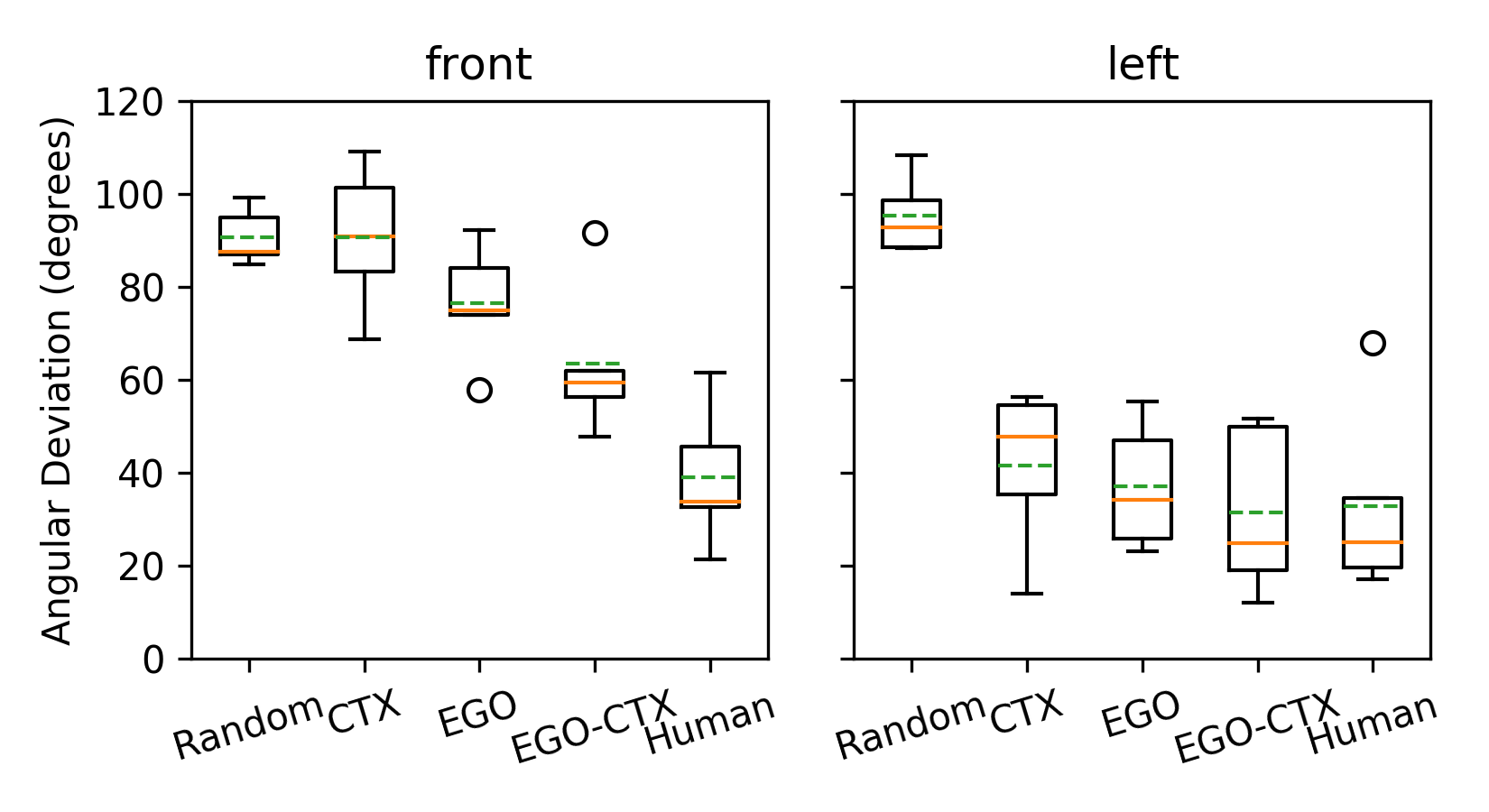}
  \caption{FoR prediction results. The solid orange line shows the median, and the dotted green line shows the mean. The circles are outliers. Lower is better.}
  \label{fig:foref_boxplot}
\end{figure}

\begin{figure}[t]
  \includegraphics[width=\linewidth]{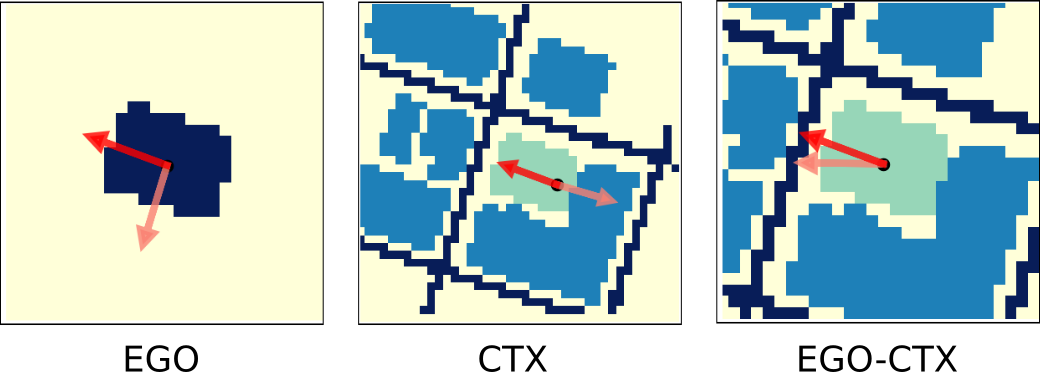}
  \caption{Visualization of FoR predictions for \emph{front}. Darker arrows indicate labeled FoR, while brighter arrows are predicted FoR.}
  \label{fig:foref_examples}
\vspace{-0.3cm}
\end{figure}

The results are shown in Figure~\ref{fig:foref_boxplot}. Each boxplot summarizes the means of baseline performance in the five cross-validation splits.
The results demonstrate that \textbf{EGO-CTX} shows generalizable behavior close to the human annotators, especially for \textbf{front}. We observe that our model is able to predict \emph{front} FoRs roughly perpendicular to streets against other buildings, while the baselines often fail to do so (Figure~\ref{fig:foref_examples}). The competitive performance of the neural network baselines in \textbf{left} indicates that for \emph{left} and \emph{right}, the FoR annotations are often absolute, i.e. independent of the context. Our model as well as baselines are limited in determining, for example, whether the speaker refers to the left side of the map (absolute FoR), or the left side of the street relative to a perceived forward direction (relative FoR).


\subsection{End-to-End Evaluation}
\label{sec:end-to-end-eval}
We randomly select 20 spatial descriptions per city.  We task the robot to search for each target object mentioned in every description separately, resulting in a total of 40 search trials per city, 200 in total. Cross-validation is employed such that for each city, the robot uses the FoR prediction model trained on the other four cities. For each step, the robot can either move or mark an object as detected. The robot can move by rotating clockwise or counterclockwise for 45 degrees, or move forward by 3 grid cells (15m). The robot receives observation through an on-board fan-shaped sensor after every move. The sensor has a field of view with an angle of 45 degrees and a varying depth of 3, 4, 5 (15m, 20m, 25m). As the field of view becomes smaller, the search task is expected to be more difficult. 
The robot receives $R_{\text{step}}=-10$ step cost for moving and $R_{\max}=+1000$ for correctly detecting the target, and $R_{\min}=-1000$ if the detection is incorrect.  The rest of the domain setup follows~\cite{wandzel2019oopomdp}.

\emph{Baselines.} \textbf{SLOOP} uses the spatial language observation model without mixture, that is, for each object, it computes the observation distribution in Eq.~(\ref{eq:sprls}) by multiplying the distributions for each spatial relation; With the same observation distribution, \textbf{SLOOP (m=2)} mixes in one distribution computed by treating all prepositions as \emph{near} with weight 0.2; Also with the same observation distribution, \textbf{SLOOP (m=4)} mixes in three additional distributions: one ignores FoR-dependent prepositions, one treating all prepositions as \emph{near}, one treating all prepositions as \emph{at}, with weights 0.25, 0.1, 0.05, respectively. The baseline \textbf{MOS (keyword)} uses a keyword-based model \cite{wandzel2019oopomdp} that assigns a uniform probability over referenced landmarks in a spatial language but does not incorporate information from spatial prepositions. Finally, \textbf{informed} and \textbf{uniform} are upper and lower bounds: for the \textbf{informed}, the agent has an initial belief that has a small Gaussian noise over the groundtruth location\footnote{The noise is necessary for object search, otherwise the task is trivial.}; \textbf{uniform} uses a uniform prior. We also report the performance with annotated spatial relations and landmarks to show search performance if the languages are parsed correctly.


For all baselines, we use an online POMDP solver, POMCP \cite{silver2010monte} but with a histogram belief representation to avoid particle depletion.
The number of simulations per planning step is 1000 for all baselines.  The discount factor is set to 0.95.  The robot is allowed to search for 200 steps per search task, since search beyond this point will earn very little discounted reward and is not efficient.

\begin{figure}[t]
  \centering
  \includegraphics[width=\linewidth]{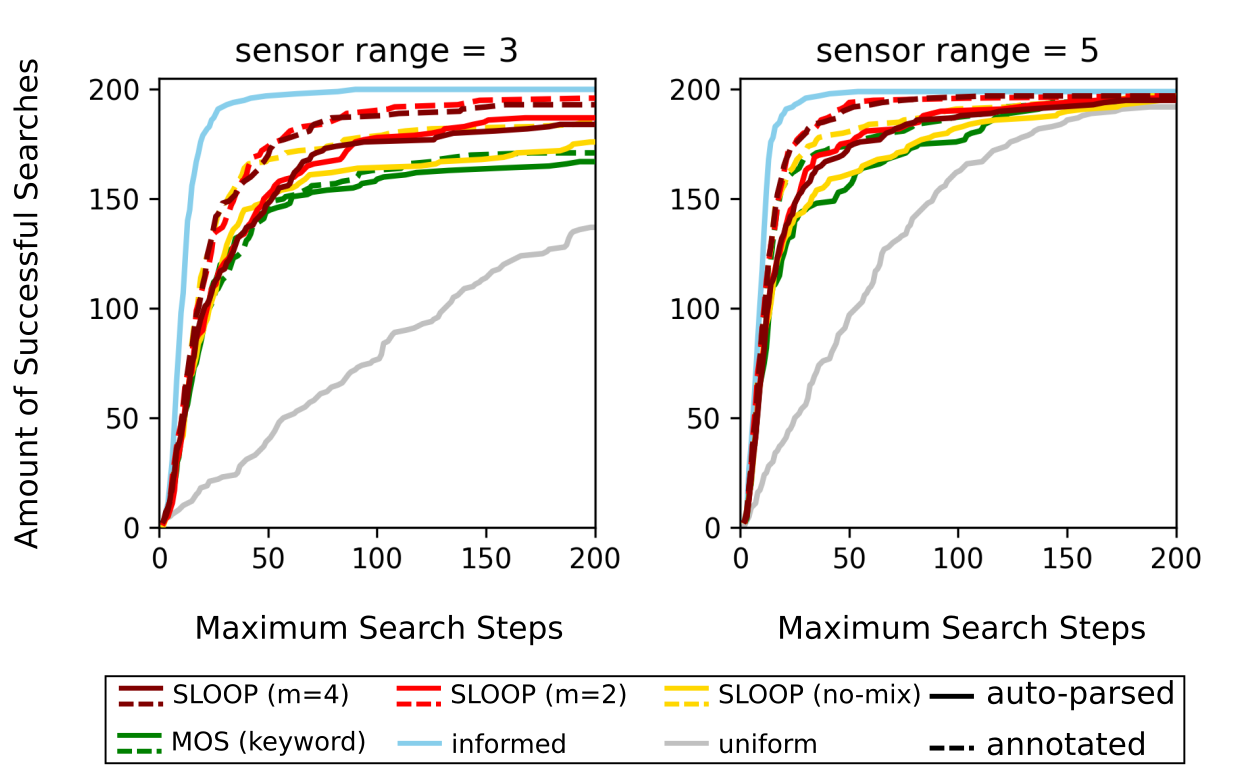}
  \caption{Number of completed search tasks as the maximum search step increases. Steeper slope indicates greater efficiency and success rate of search.}
  \label{fig:detections}
  \vspace{-0.3cm}
\end{figure}

\begin{figure}[t]
  \centering
  \includegraphics[width=\linewidth]{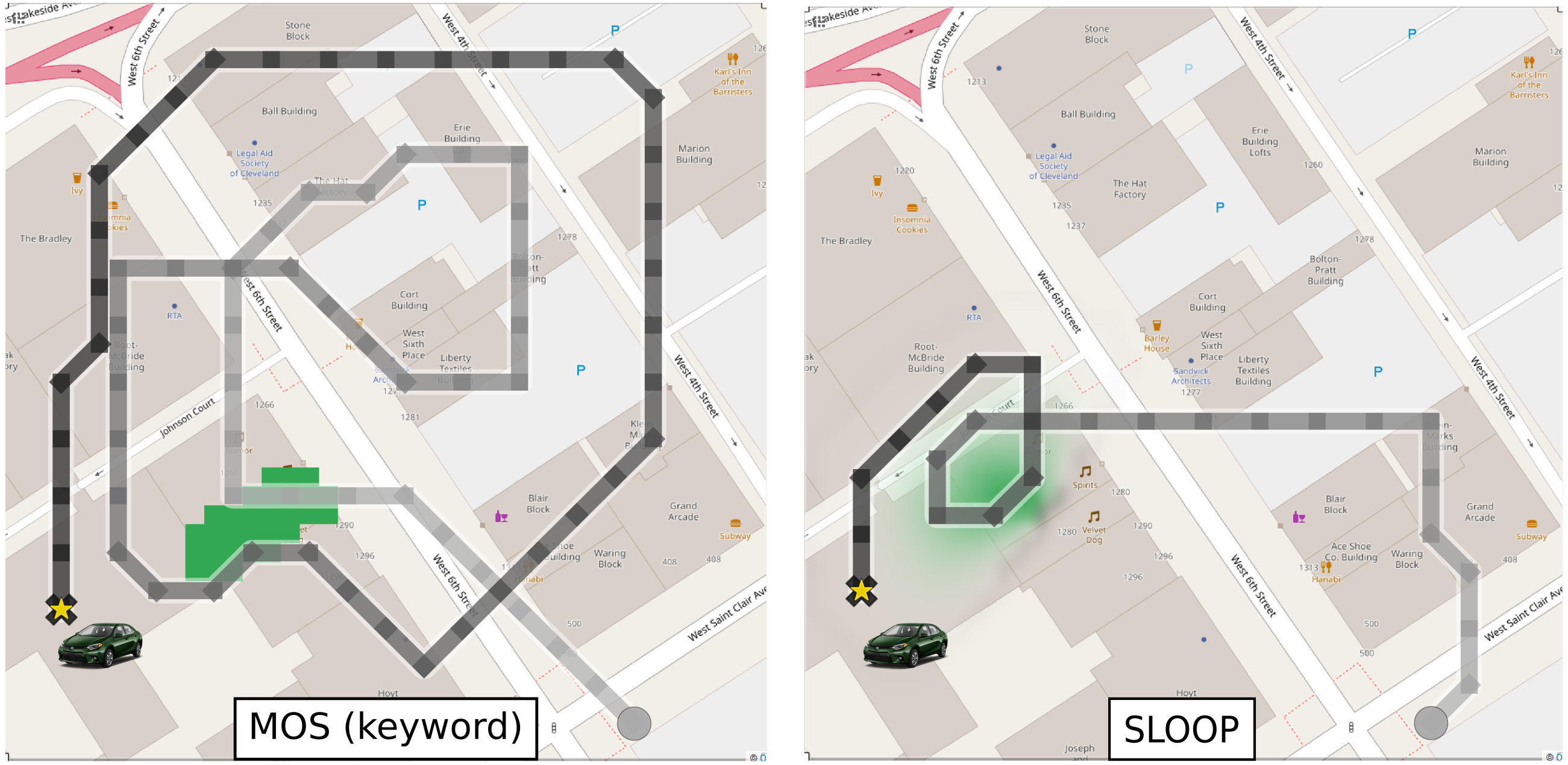}
  \caption{Example object search trial for description ``the green toyota is
    \emph{behind} velvet dog'' from AMT.  The green region shows the
    distribution over the object location after interpreting the description.
    Our method enables probabilistic interpretation of the spatial language
    leading to more efficient search strategy.}
  \label{fig:osmexample}
  \vspace{-0.2cm}
\end{figure}


\begin{table}[tb]
  \resizebox{\linewidth}{!}{%
    \begin{tabular}{rr|rrr}
      \toprule
      spatial     &  No.            & MOS (keyword) \cite{wandzel2019oopomdp}                          & SLOOP                         & SLOOP (m=4)\\              
      preposition &  trials    & annodated                                              & annotated                     & annotated\\                
      \midrule                                                                                                                                       %
      on	& 59	       & 200.65 (78.06)	                                        & 267.00 (76.45)	        &  \textbf{290.05 (70.51)}\\                          
      at	& 58	       & 179.42 (81.46)	                                        & 237.36 (81.03)	        &  \textbf{238.24 (80.50)} \\                         
      near	& 35	       & 97.64 (135.39)	                                        & \textbf{280.69 (109.00)}	&  249.35 (113.60) \\                         
      between	& 25	       & 21.48 (116.93)	                                        & 172.93 (143.77)	&  \textbf{175.59 (136.71)}  \\                         
      in	& 22	       & 302.05 (151.42)	                                & \textbf{398.88 (119.32)}	&  307.45 (141.67) \\                         
      \midrule
      north	& 9	       & 222.28 (291.13)	                                & 201.88 (296.49)	        & \textbf{365.14 (246.05)}\\
      southeast	& 7	       & 306.77 (341.43)	                                & 553.82 (174.04)	        & \textbf{549.43 (165.83)}\\
      southwest	& 7	       & -75.67 (205.98)	                                & \textbf{1.37 (271.46)}	                & -27.63 (281.89)\\
      east	& 6	       & 56.57 (337.58)	                                        & 290.68 (303.53)	        & \textbf{439.99 (276.85)}\\
      northwest	& 6	       & \textbf{385.41 (320.82)}	                                & 43.57 (282.88)	        & -1.82 (256.71) \\
      south	& 6	       & 79.12 (289.54)	                                        & 310.29 (410.04)	        & \textbf{494.26 (161.60)}\\

      west	& 4	       & -160.91 (188.02)	                                & 234.93 (587.57)	        & \textbf{327.13 (245.28)}\\

      northeast	& 2	       & -167.99 (660.76)	                                & 206.42 (138.93)	        & \textbf{213.17 (977.62)}\\
      \midrule
      front	& 25	       & \textbf{246.96 (142.45)}	                        & 168.91 (150.41)	        & 160.55 (136.88)  \\                         
      behind    & 8            & 128.47 (356.25)	                                & 101.20 (333.38)	        & \textbf{140.92 (333.61)}\\
      right	& 4	       & 19.75 (697.88)	                                        & 160.14 (601.54)	        & \textbf{336.00 (725.84)}\\
      left	& 3	       & \textbf{247.35 (363.32)}	                        & 192.93 (393.75)	        & 231.75 (330.33)\\
      \midrule
      front (good) & 15	       & 255.85 (210.46)		                        & \textbf{421.83 (143.65)}	& 222.67 (264.50) \\                          
      behind (good) & 6	       & 145.26 (489.55)		                        & 207.58 (430.52)		& \textbf{359.80 (753.15)}     \\        
      front (bad)  & 10        & \textbf{281.04 (226.47)}				& -208.92 (11.39)	        & 93.80 (176.11)             \\              
      behind (bad) & 2	       & \textbf{78.11 (3771.84)}	                        & -217.95 (23.53)	        & -77.97 (223.71)                \\           

      \bottomrule
\end{tabular}
}
\caption{Mean (95\% CI) of discounted cumulative reward for different prepositions
  evaluated on language descriptions with annotated spatial relations. The value with highest mean per row is bolded.
  }
\label{tab:prepositions}

\end{table}

\begin{table}[ht]
\resizebox{\linewidth}{!}{%
\begin{tabular}{rr|rrr}
\toprule
  No. spatial~      &   No.~   & MOS (keyword) \cite{wandzel2019oopomdp}    & SLOOP                         & SLOOP (m=4)\\              
  prepositions      & trials   & annotated                                  & annotated                     & annotated\\
\midrule
1	          &    100           & 234.32 (72.64)	                        & \textbf{320.91 (64.23)}		& 289.34 (66.42)\\
2                 &    83            & 179.18 (68.60)	                        & 264.08 (62.39)		& \textbf{286.19 (60.83)}\\
3	          &    14            & 26.96 (165.98)	                        & 115.44 (202.42)		& \textbf{215.30 (200.99)}\\
\bottomrule
\end{tabular}
}
\caption{Mean (95\% CI) of discounted cumulative reward for completed search tasks as language complexity (number of spatial relations) increases.}
\label{tab:numrels}
\vspace{-1em}
\end{table}

\emph{Results.} We evaluate the effectiveness and efficiency of the search by the amount of search tasks the robot completed (i.e. successfully found the target) under a given limit of search steps (ranging from 1 to 200).
Results are shown in Figure~\ref{fig:detections}. The results show that using spatial language with SLOOP outperforms the keyword-based approach in MOS. The gain in the discounted reward is statistically significant over all sensor ranges comparing \textbf{SLOOP} with \textbf{MOS (keyword)}, and for sensor range 3 comparing the annotated versions. We observe that using mixture models for spatial language improves search efficiency and success rate over \textbf{SLOOP}. We observe improvement when the system is given annotated spatial relations. This suggests the important role of language parsing for spatial language understanding.
Figure~\ref{fig:osmexample} shows a trial comparing \textbf{SLOOP} and \textbf{MOS (keyword)}.

\begin{figure}[tb]
  \centering
  \includegraphics[width=\linewidth]{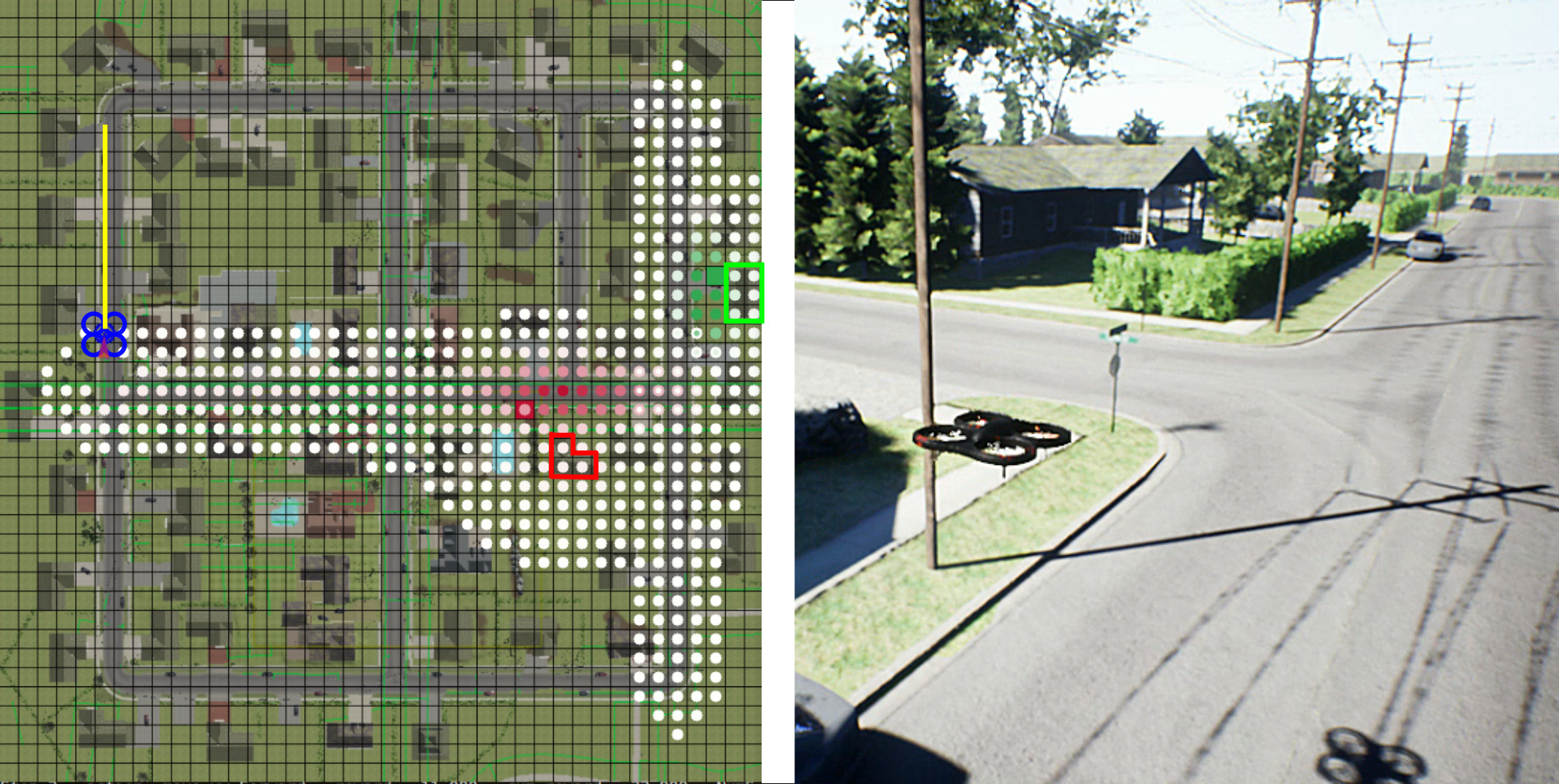}
  \caption{Example trial from AirSim demonstration. Given spatial language description: \emph{The red car is by the side of Main street in front of Zoey's House (red), while the green car is within Annie's House on the right side of East street (green).} Left: belief over two target objects (red and green car). Right: screenshot from AirSim. Refer to the supplementary video for more details.}

  \label{fig:airsimexample}
\vspace{-0.1cm}
\end{figure}

We analyze the performance with respect to different spatial prepositions. We report results for annotated languages as they reflect the performance obtained if the prepositions are correctly identified. Results for the smallest sensor range of 3 is shown in Table~\ref{tab:prepositions}.
\textbf{SLOOP} outperforms the baseline for the majority of prepositions.
For prepositions \emph{front}, \emph{behind}, \emph{left}, and \emph{right}, our investigation shows the performance of \textbf{SLOOP} polarizes where trials with ``good'' FoR (i.e. ones in the correct direction towards the true target location) leads to a much greater performance than the counterpart (``bad'' FoR). Yet, \textbf{MOS (keyword)} is not subject to such polarization and the target often appears close to the landmark for these prepositions. We observe that \textbf{SLOOP (m=4)} using mixture is able to consistently improve the reward for most of the prepositions, indicating the benefit of modeling multiple interpretations of the spatial language.

Finally, we analyze the relationship between the performance and varying complexity of the language description, indicated by the number of spatial relations used to describe the target location.  Again, we used annotated languages for this experiment for the smallest sensor range of 3. Results in Table~\ref{tab:numrels} indicate that understanding spatial language can benefit search performance, with a wider gain as the number of spatial relations increases.
Again, using mixture model in \textbf{SLOOP (m=4)} improves the performance even more.

\subsection{Demonstration on AirSim}
We implemented \textbf{SLOOP (m=4)} on AirSim \cite{airsim2017fsr}, a realistic drone simulator built on top of Unreal Engine \cite{unrealengine}. Similar to our evaluation in OpenStreetMap, we discretize the map into 41$\times$41 grid cells. We use the same fan-shaped model of on-board sensor as in OpenStreetMap. As mentioned in Sec.~\ref{sec:setting}, sensor observations are synthetic, based on the ground-truth state. Additionally, although the underlying localization and control is continuous, the drone plans discrete navigation actions (move forward, rotate left 90$^{\circ}$, rotate right 90$^{\circ}$). We annotated landmarks (houses and streets) in the scene on the 2D grid map. Houses with heights greater than flight height are subject to collision and results in a large penalty reward (-1000). Checking for collision in the POMDP model for this domain helped prevent such behavior during planning. We found that the FoR prediction model trained on OpenStreetMap generalizes to this domain, consistently producing reasonable FoR predictions for \emph{front} and \emph{behind}. This shows the benefit of using synthetic images of top-down street maps. The drone is able to plan actions to search back and forth to find the object, despite given inexact spatial language description. Please refer to the video demo on our project for the examples shown in Fig.~\ref{fig:firstfig} and \ref{fig:airsimexample}.

\section{Conclusions}
\label{sec:conclusion}
This paper first presents a formalism for integrating spatial language into the POMDP belief state as an observation, then a convolutional neural network for FoR prediction shown to generalize well to new cities. Simulation experiments show that our system significantly improves object search efficiency and effectiveness in city-scale domains
through understanding spatial language. For future work, we plan to investigate compositionality in spatial language for partially observable domains.

\section*{Acknowledgements}
We thank Thao Nguyen and Eric Rosen for valuable feedback on initial revisions. This work was supported by the National Science Foundation under grant number
IIS1652561, the US Army under grant number W911NF1920145, DARPA under grant
number HR00111990064, Echo Labs, STRAC Institute, and Hyundai.

\bibliography{references}

\newpage
\onecolumn

\input{suppl.tex}

\end{document}

%% file: suppl.tex
\section{Appendix}

\subsection{Derivation of Spatial Language Observation Model}

Here we provide the derivation for Eq.~(\ref{eq:sprls}).
Using the definition of $o_{\sprl_i}$, 
\begin{align}
\Pr(o_{\sprl_i} | s_i', \mathcal{M})&=\Pr(f_i, r_1,\gamma_1,\cdots,r_L,\gamma_L | s_i', \mathcal{M})\\
&=\Pr(r_1,\cdots,r_L | \gamma_1,\cdots,\gamma_L,f_i, s_i', \mathcal{M})
\times\Pr(\gamma_1,\cdots,\gamma_L, f_i| s_i', \mathcal{M})\label{eq:por1}\\
&=\frac{1}{Z}\prod_{j=1}^L\Pr(r_j | \gamma_j,f_i, s_i', \mathcal{M})\label{eq:por2}
\end{align}
The first term in (\ref{eq:por1}) is factored by individual spatial relations, because each $r_j$ is a predicate that, by definition, involves only the figure $f_i$ and the ground $\gamma_j$, therefore it is conditionally independent of all other relations and grounds given $f_i$, its location $s_i'$, and the landmark $\gamma_j$ and its features contained in $\mathcal{M}$. Because the robot has no prior knowledge regarding the human observer's language use,\footnote{In general, the human observer may produce spatial language that mentions arbitrary landmarks and figures whether they make sense or not.}  the second term in (\ref{eq:por1}) is uniform with probability $1/Z$ where $Z$ is the constant size of the support for $\gamma_1,\cdots,\gamma_L,f_i$. This constant can be canceled out during POMDP belief update upon receiving the spatial language observation, using the belief update formula in  Section~\ref{sec:prelim}. We omit this constant in Eq.~(\ref{eq:sprls}).

For predicates such as \emph{behind}, its truth value depends on the relative FoR imposed by the human observer who knows the target location. Denote the FoR vector corresponding to $r_{j}$ as a random variable $\Psi_{j}$ that distributes according to the indicator function $\Pr(\Psi_{j}=\psi_{j})=\mathbbm{1}(\psi_{j}=\psi_{j}^*)$, where $\psi_{j}^*$ is the one imposed by the human. Then regarding $\Pr(r_{j}|\gamma_j,f_i,s_i',\mathcal{M})$, we can sum out $\Psi_j$:
\begin{align}
  \Pr(r_{j}|\gamma_j,f_i,s_i',\mathcal{M})&=\frac{\sum_{\psi_j}\Pr(r_{j},\gamma_j,f_i,s_i',\mathcal{M}|\psi_j)\Pr(\psi_j)}{\Pr(\gamma_j,f_i,s_i',\mathcal{M})}\\
  \intertext{Since the distribution for $\Psi_j$ is an indicator function,}
                                          &=\frac{\Pr(r_{j},\gamma_j,f_i,s_i',\mathcal{M}|\psi_j^*)}{\Pr(\gamma_j,f_i,s_i',\mathcal{M})}\label{eq:por3}\\
  \intertext{By the law of total probability,}
                                          &=\frac{\Pr(r_j|\gamma_j,\psi_j^*,f_i,s_i',\mathcal{M})\Pr(\gamma_j,f_i|s_i',\mathcal{M},\psi_j^*)\Pr(s_i',\mathcal{M}|\psi_j^*)}{\Pr(\gamma_j,f_i|s_i',\mathcal{M})\Pr(s_i',\mathcal{M})}\\
  \intertext{Using the fact that $s_i',\mathcal{M}$ is independent of $\psi_j^*$,}
                                          &=\frac{\Pr(r_j|\gamma_j,\psi_j^*,f_i,s_i',\mathcal{M})\Pr(\gamma_j,f_i|s_i',\mathcal{M},\psi_j^*)\Pr(s_i',\mathcal{M})}{\Pr(\gamma_j,f_i|s_i',\mathcal{M})\Pr(s_i',\mathcal{M})}\\
  \intertext{Canceling out $\Pr(s_i',\mathcal{M})$,}
                                          &=\frac{\Pr(r_j|\gamma_j,\psi_j^*,f_i,s_i',\mathcal{M})\Pr(\gamma_j,f_i|s_i',\mathcal{M},\psi_j^*)}{\Pr(\gamma_j,f_i|s_i',\mathcal{M})}\\
  \intertext{Similar to (\ref{eq:por1})-(\ref{eq:por2}), $\Pr(\gamma_j,f_i|s_i',\mathcal{M},\psi_j^*)$ and $\Pr(\gamma_j,f_i|s_i',\mathcal{M})$ are uniform with the same support. Canceling them out,}
&=\Pr(r_j | \gamma_j,\psi_j^*,f_i, s_i', \mathcal{M})\label{eq:por4}
\end{align}

\subsection{Data Collection Details}

\subsubsection{Amazon Mechanical Turk Questionnaire}
\begin{figure}[!]
\begin{subfigure}{\textwidth}
  \centering
  \includegraphics[width=0.65\linewidth]{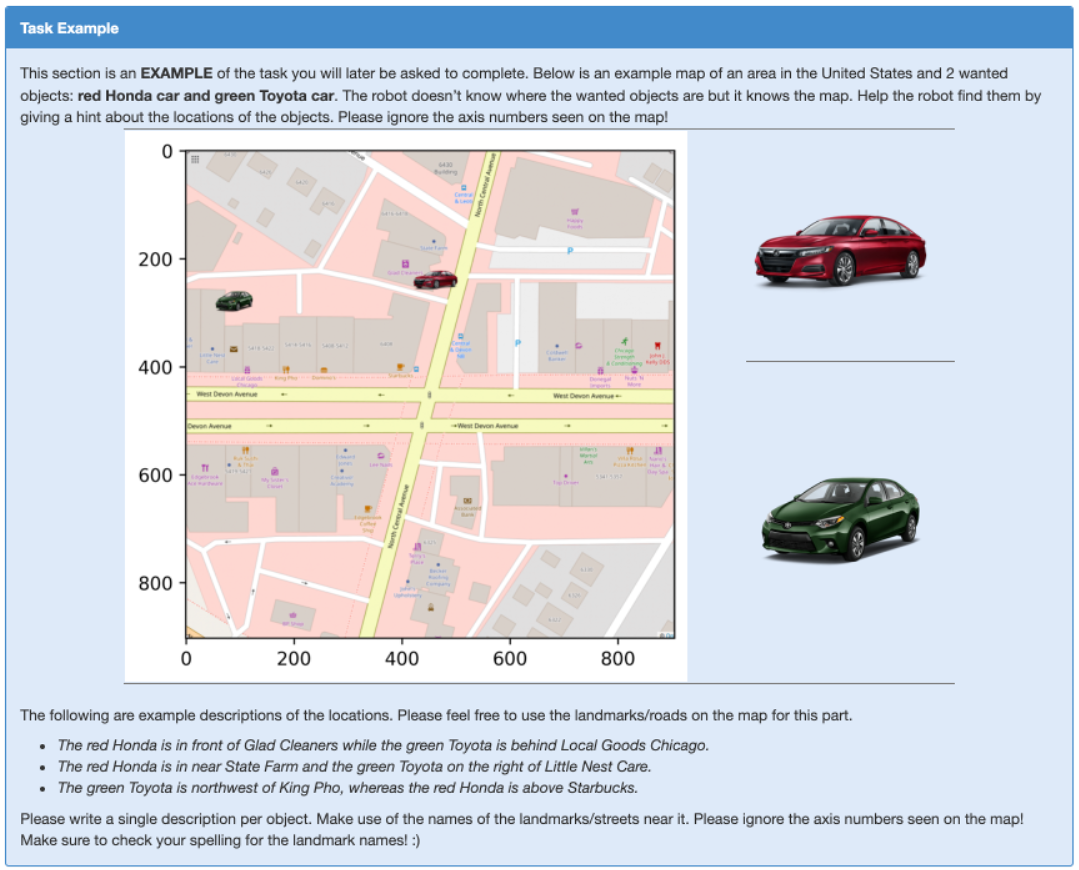}
\end{subfigure}
\begin{subfigure}{\textwidth}
  \centering
  \includegraphics[width=0.65\linewidth]{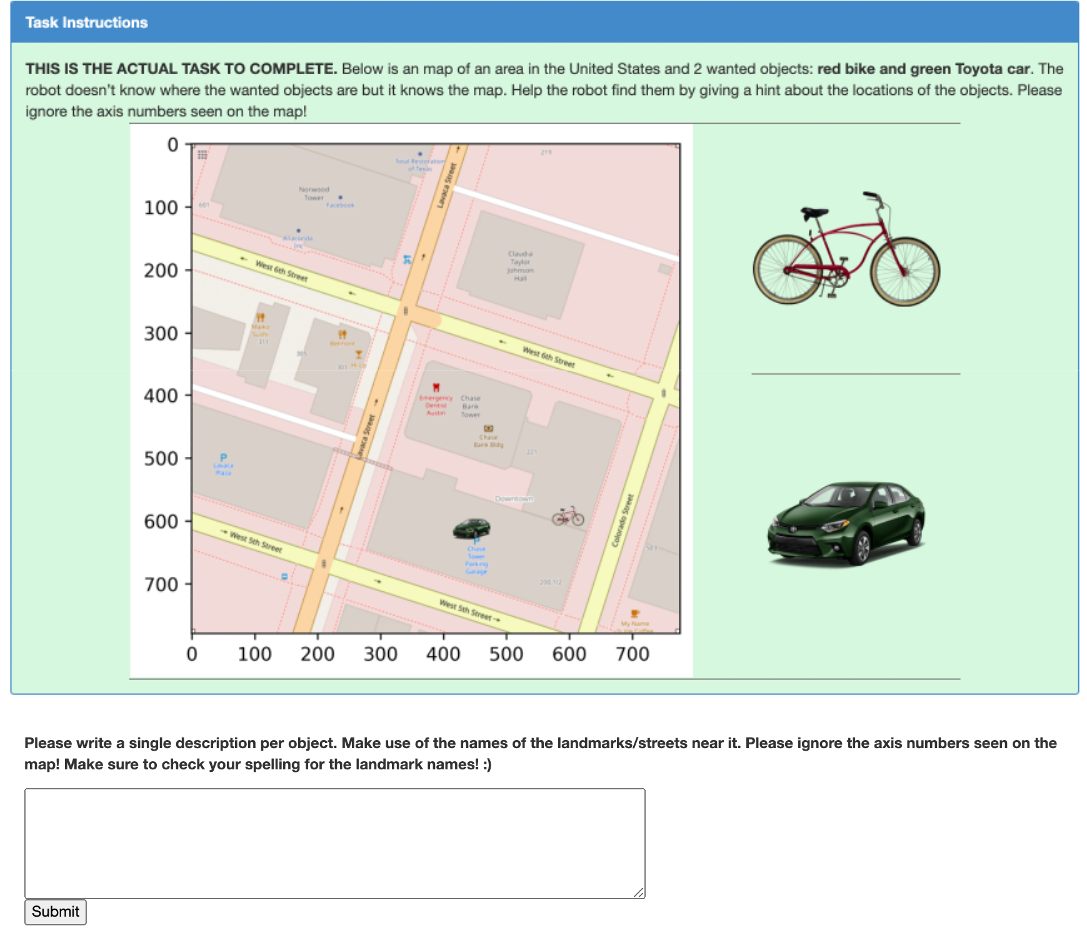}
\end{subfigure}
\caption{AMT Questionnaire Screenshot. Top: an example task shown to the AMT workers prior to the actual task that doesn't vary from prompt to prompt. Bottom: the actual task shown to the AMT workers. The objects and the locations change in every prompt and are all unique.}
\label{fig:amt-questionnaire}
\end{figure}

As described in Section~\ref{sec:data_collect}, we collect a variety of spatial language descriptions from five cities. We randomly generate 10 unique configurations of two object locations for every pair of object symbols from \{\texttt{RedBike}, \texttt{RedCar}, \texttt{RedCar}\}. Each configuration is used to obtain language descriptions from up to eleven different workers.
By showing a picture of the objects placed on the map screenshot, we prompt AMT workers to describe the location of the target objects. We first show an example task as shown in Fig~\ref{fig:amt-questionnaire} (top). Then we prompt them with the actual task and a text box to submit their response, as shown in Fig~\ref{fig:amt-questionnaire} (bottom). Note that we specify that the robot does not know where the target objects are, but that it knows the buildings, streets and other landmarks available on the map. We encourage them to use the information available on the map in their description.


\subsubsection{Distribution of Collected Predicates}
\begin{figure}[H]
\centering
\includegraphics[width=0.96\linewidth]{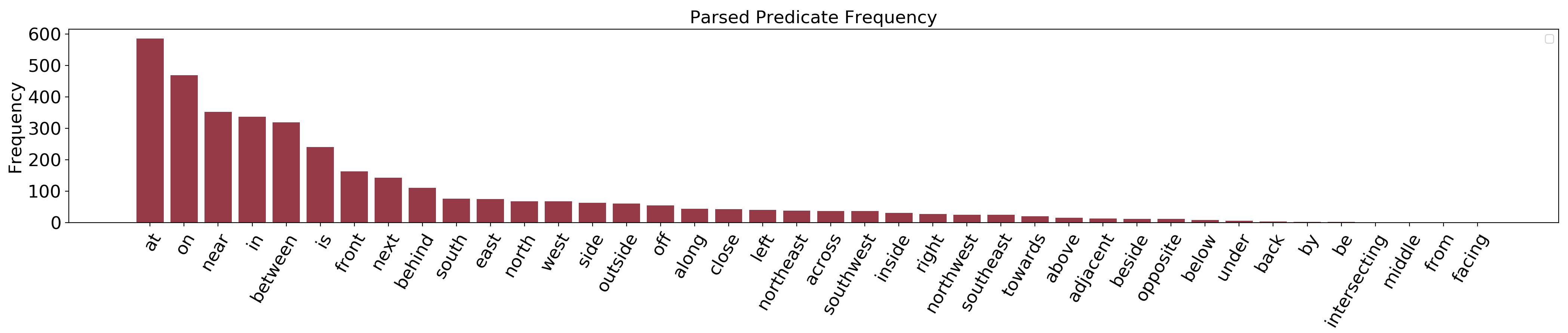}
\caption{Distribution of collected parsable predicates sorted from most frequent to least.}
\label{fig:pred-dist}
\end{figure}

Each description is parsed using our pipeline described in Section~\ref{sec:spatial_info_extraction}. 1,521 out of 1,650 gathered descriptions were successfully parsed; meaning at least one spatial relation was extracted from the sentence. The distribution of all spatial predicates are shown in Fig~\ref{fig:pred-dist}. Note that we include the word ``is'' on the list since it often appears in  ``is in'' or ``is at'', yet the parser sometimes skip the word after it due to an artifact. We excluded language descriptions parsed with such artifact from the ones used in the end-to-end object search evaluation.


\begin{table}[H]
\centering
\begin{tabular}{rr}
\toprule
Spatial preposition & \# of FoR annotations \\
\midrule
front            & 121              \\
behind           & 54               \\
left             & 51               \\
right            & 47               \\
\bottomrule
\end{tabular}
\caption{Number of FoR annotations per spatial relation.}
\label{tab:number-for-annot}
\end{table}

\subsubsection{FoR Annotation}
\begin{figure}[H]
  \centering
  \includegraphics[width=0.67\linewidth]{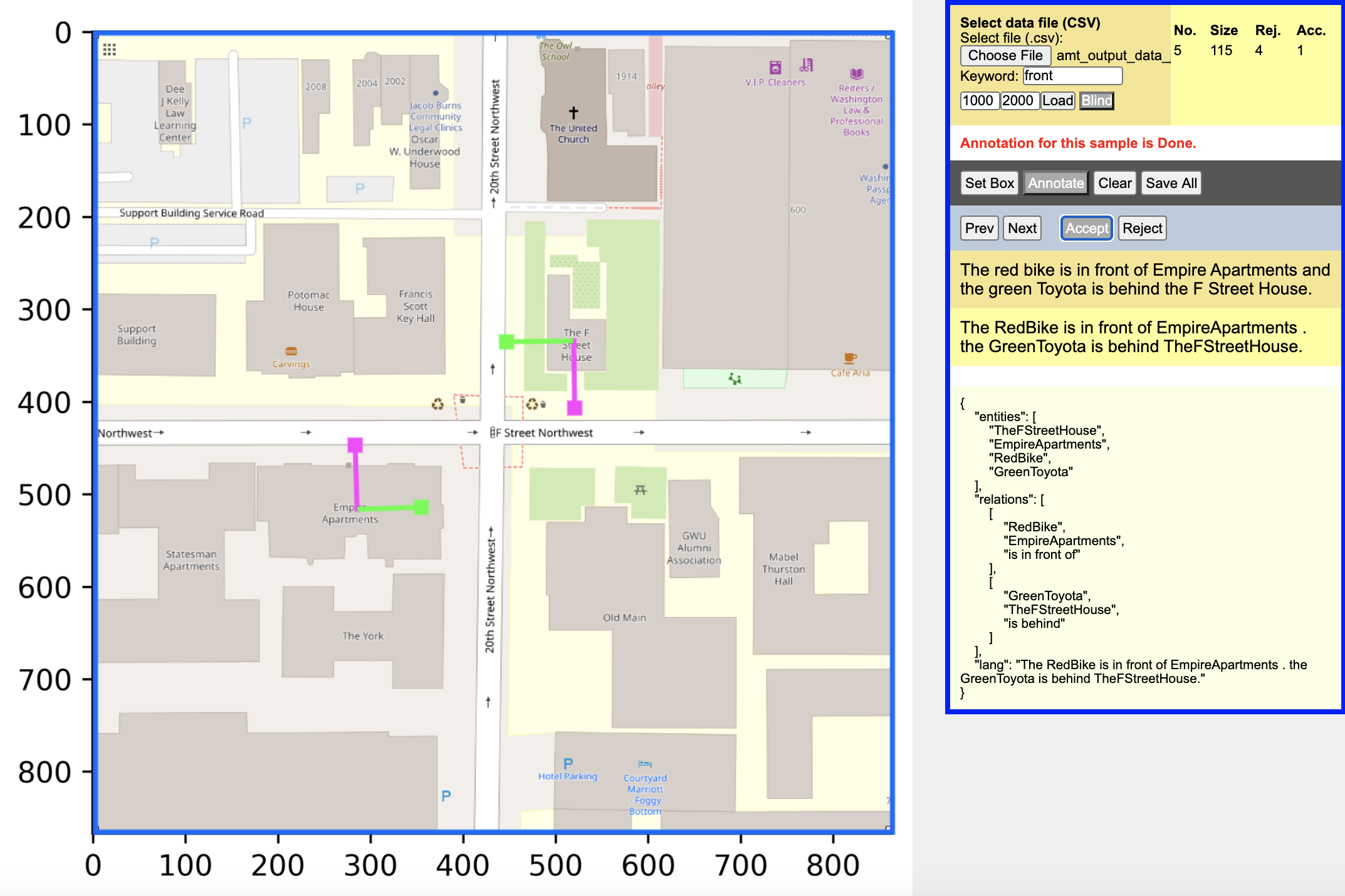}
\caption{The FoR annotator GUI interface. The FoR consists of a front (purple) and right (green) vector. The target objects are not shown, and the annotator only has access to the spatial language description, and the map image. This mimics the situation faced by the robot in our task.}
\label{fig:annotator-gui}
\end{figure}

To collect the FoR annotation, we create our custom annotation tool. The FoR consists of a front (purple) and right (green) vector that show how the speaker considers the direction of the ground according to their given spatial description. The interface (Fig~\ref{fig:annotator-gui}) works as follows: (1) The annotator first clicks the ``Annotate'' button. (2) The interface prompts the annotator the language phrase corresponding to a spatial relation to be annotated (e.g. ``\texttt{RedBike} \texttt{is in front of} \texttt{EmpireApartments}''), which is composed using the parsed $(f,r,\gamma)$ tuple. (3) to annotate an FoR, the annotator clicks on the map as the origin of the FoR, and then clicks on another point on the map as the end point of the \emph{front} vector. The vector for \emph{right} is automatically computed to be 90 degrees clockwise with respect to the \emph{front} vector. (4) After annotating one FoR, the annotator clicks ``Next'' to move on, and the process starts over again from step (2). In total we have 273 FoR annotations. Table~\ref{tab:number-for-annot} shows the amount of annotation per spatial relation. You can download the dataset by visiting the website linked in the footnote on the first page.


